\documentclass[runningheads,10pt]{llncs}
\usepackage{graphicx}
\usepackage{amsmath,amssymb} 
\usepackage{color}
\usepackage{xspace}
\usepackage{lettrine}
\usepackage{array}
\usepackage[width=122mm,left=12mm,paperwidth=146mm,height=193mm,top=12mm,paperheight=217mm]{geometry}
\begin{document}

\pagestyle{headings}
\mainmatter
\def\ECCV14SubNumber{411}  

\title{Orientation covariant aggregation of local descriptors with embeddings} 

\titlerunning{Orientation covariant aggregation of local descriptors with embeddings}

\authorrunning{Giorgos Tolias, Teddy Furon \& Herv\'e J\'egou}

\author{Giorgos Tolias, Teddy Furon \& Herv\'e J\'egou}
\institute{Inria}

\newcommand{\real}{\mathbb{R}}
\newcommand{\tran}{^{\mathrm{T}}}
\def\alphab{\boldsymbol{\alpha}}
\def\xb{\mathbf{x}}
\def\yb{\mathbf{y}}
\def\phib{{\varphi}}

\def\Xset{{\mathcal X}}
\def\Yset{{\mathcal Y}}
\def\Xf{{\mathcal X}^\star}
\def\Yf{{\mathcal Y}^\star}
\def\Kf{K^\star}

\def\Xb{\mathbf{X}}
\def\Yb{\mathbf{Y}}
\def\Xbf{\mathbf{X}^\star}
\def\Ybf{\mathbf{Y}^\star}

\def\Xsetrot{\breve{\mathcal X}}
\def\Ysetrot{\breve{\mathcal Y}}
\def\Xrot{\mathbf{\breve{X}}^\star}
\def\Yrot{\mathbf{\breve{Y}}^\star}

\def\VM{\mathsf{VM}}

\def\etal{\emph{et al.}\xspace}
\def\ie{\emph{i.e.}\xspace}

\def\nssp{\hspace{-1pt}}
\def\sssp{\hspace{2pt}}
\def\ssp{\hspace{3pt}}
\def\msp{\hspace{5pt}}
\def\bsp{\hspace{12pt}}

\maketitle

\begin{abstract}
Image search systems based on local descriptors typically achieve orientation invariance by aligning the patches on their dominant orientations. Albeit successful, this choice introduces too much invariance because it does not guarantee that the patches are rotated consistently. 

This paper introduces an aggregation strategy of local descriptors that achieves this covariance property by jointly encoding the angle in the aggregation stage in a continuous manner. It is combined with an efficient monomial embedding to provide a codebook-free method to aggregate local descriptors into a single vector representation. 

Our strategy is also compatible and employed with several popular encoding methods, in particular bag-of-words, VLAD and the Fisher vector. Our geometric-aware aggregation strategy is effective for image search, as shown by experiments performed on standard benchmarks for image and particular object retrieval, namely Holidays and Oxford buildings.
\end{abstract}

\section{Introduction}
\label{sec:introduction}

\lettrine{T}{his} paper considers the problem of particular image or particular object retrieval. This subject has received a sustained attention over the last decade. Many of the recent works employ local descriptors such as SIFT~\cite{L04} or variants~\cite{BETV08} for the low-level description of the images. In particular, approaches derived from the bag-of-visual-words framework~\cite{SZ03} are especially successful to solve problems like recognizing buildings. They are typically combined with spatial verification~\cite{PCISZ07} or other re-ranking strategies such as query expansion~\cite{CPSIZ07}. 

Our objective is to improve the quality of the first retrieval stage, before any re-ranking is performed. This is critical when considering large datasets, as re-ranking methods depend on the quality of the initial short-list, which typically consists of a few hundred images.
The initial stage is improved by better matching rules, for instance with Hamming embedding~\cite{JDS10a}, by learning a fine vocabulary~\cite{MPCM10}, or weighting the distances~\cite{JDS09a,TAJ13}. In addition to the SIFT, it is useful to employ some geometrical information associated with the region of interest~\cite{JDS10a}. All these approaches rely on matching individual descriptors and therefore store some data on a per descriptor basis. Moreover, the quantization of the query's descriptors on a relatively large vocabulary causes delays. 

Recently, very short yet effective representations have been proposed based on alternative encoding strategies, such as local linear coding~\cite{WYY10}, the Fisher vector~\cite{PD07} or VLAD~\cite{JPDSPS11}. Most of these representations have been proposed first for image classification, yet also offer very effective properties in the context of extremely large-scale image search. A feature of utmost importance is that they offer vector representations compatible with cosine similarity. The representation can then be effectively binarized~\cite{PLSP10} with cosine sketches, such as those proposed by Charikar~\cite{C02} (\emph{a.k.a.} LSH), or aggressively compressed with principal component dimensionality reduction (PCA) to very short vectors. Product quantization~\cite{JDS11} is another example achieving a very compact representation of a few dozens to hundreds bytes and an efficient search because the comparison is done directly in the compressed domain. 

This paper focuses on such short- and mid-sized vector representations of images. Our objective is to exploit some geometrical information associated with the regions of interest. A popular work in this context is the spatial pyramid kernel~\cite{LSP06}, which is widely adopted for image classification. However, it is ineffective for particular image and object retrieval as the grid is too rigid and the resulting representation is not invariant enough, as shown by Douze~\etal~\cite{DJSAS09}. 

Here, we aim at incorporating some relative angle information to ensure that the patches are consistently rotated. In other terms, we want to achieve a covariant property similar to that offered by Weak Geometry Consistency (WGC)~\cite{JDS10a}, but directly implemented in the coding stage of image vector representations like Fisher, or VLAD. Some recent works in classification~\cite{CVIU13} and image search~\cite{ZJG13} consider a similar objective. They suffer from several shortcomings. In particular, they simply quantize the angle and use it as a pooling variable. Moreover the encoding of a rough approximation of the angles is not straightforwardly compatible with generic match kernels.

In contrast, we achieve the covariant property for any method provided that it can be written as a match kernel. This holds for the Fisher vector, LLC, bag-of-words and efficient match kernels listed in~\cite{BS09}. Our method is inspired by the kernel descriptor of Bo \etal~\cite{BRF10}, from which we borrow the idea of angle kernelization. Our method however departs from this work in several ways. First, we are interested in aggregating local descriptors to produce a vector image representation, whereas they construct new local descriptors. Second, we do not encode the gradient orientation but the dominant orientation of the region of interest jointly with the corresponding SIFT descriptor, in order to achieve the covariant property of the local patches.
Finally, we rely on explicit feature maps~\cite{VZ12} to encode the angle, which provides a much better approximation than efficient match kernel for a given number of components. 

This paper is organized as follows. Section~\ref{sec:related} introduces notation and discusses some important related works more in details. Our approach is presented in Section~\ref{sec:method} and evaluated in Section~\ref{sec:experiments} on several popular benchmarks for image search, namely Oxford5k~\cite{PCISZ07}, Oxford105k and Inria Holidays~\cite{JDS08}. These experiments show that our approach gives a significant improvement over the state of the art on image search with vector representations. Importantly, we achieve competitive results by combining our approach with monomial embeddings, \ie, with a \emph{codebook-free} approach, as opposed to coding approaches like VLAD. 

\section{Preliminaries: match kernels and monomial embeddings}
\label{sec:related}

We consider the context of match kernels. An image is typically described by a set  of local descriptors $\Xset=\{\xb_1,\dots,\xb_i, \dots\},\ \xb_i \in \real^d, \|\xb_i\|=1.$  
Similar to other works~\cite{L05,BS09,JDS10a}, two images described by $\Xset$ and $\Yset$ are compared with a match kernel $K$ of the form
\begin{equation}
K(\Xset,\Yset) =
\beta(\Xset) \beta(\Yset)
\sum_{\xb \in \Xset} \sum_{\yb \in \Yset} k(\xb,\yb),
\end{equation}
where $k$ is referred to as the local kernel and where the proportionality factor~$\beta$ ensures that $K(\Xset,\Xset)=K(\Yset,\Yset)=1$. A typical way to obtain such a kernel is to map the vectors $\xb$ to a higher-dimensional space with a function $\phib: \real^d \rightarrow\real^D$, such that the inner product similarity evaluates the local kernel 
$k(\xb,\yb)=\langle \phib(\xb)|\phib(\yb) \rangle$. This approach then represents a set of local descriptors by a single vector 
\begin{equation}
\Xb = \beta(\Xset) \sum_{\xb \in \Xset} \phib(\xb_i), \quad \quad \quad \quad \text{(such that $\| \Xb \|=1$)}
\end{equation}
because the match kernel is computed with a simple inner product as
\begin{equation}
K(\Xset,\Yset) = \beta(\Xset) \beta(\Yset) \sum_{\xb \in \Xset} \sum_{\yb \in \Yset} \langle \phib(\xb)|\phib(\yb) \rangle
= \langle \Xb | \Yb \rangle. 
\label{equ:matchvector}
\end{equation}

This framework encompasses many approaches such as bag-of-words~\cite{SZ03,CDFWB04}, LLC~\cite{WYY10}, Fisher vector~\cite{PD07}, VLAD~\cite{JPDSPS11}, or VLAT~\cite{PG13}. Note that some non-linear processing, such as power-law component-wise normalization~\cite{JDS09a,PSM10}, is often applied to the resulting vector. 
A desirable property of $k$ is to have $k(\xb,\yb)\approx 0$ for unrelated features, so that they do not interfere with the measurements between the true matches. It is somehow satisfied with the classical inner product $k(\xb,\yb)=\langle \xb|\yb \rangle$. 
Several authors~\cite{L05,PG13,TAJ13} propose to increase the contrast between related and unrelated features with a monomial match kernel of degree $p$ of the form
\begin{equation}
K(\Xset,\Yset) =
\beta(\Xset) \beta(\Yset)
\sum_{\xb \in \Xset} \sum_{\yb \in \Yset} \langle \xb|\yb \rangle^p. 
\label{equ:mono}
\end{equation}
All monomial (and polynomial) embeddings admit exact finite-dimensional feature maps whose length rapidly increases with degree $p$ (in ${\mathcal O}(d^p/p!))$.
The order $p=2$ has already demonstrated some benefit, for instance recently for semantic segmentation~\cite{CBS12} or in image classification~\cite{PG13}.
In this case, the kernel is equivalent to comparing the set of features based on their covariance matrix~\cite{PG13}. Equivalently, by observing that some components are identical, we can define the embedding $\phib_{2}:\real^d \rightarrow \real^{d(d+1)/2}$ mapping $\xb={[x_1,\dots,x_d]^\top} $ to
\begin{equation}
\phib_2(\xb) = [x_1^2,\dots,x_d^2,x_1 x_2 \sqrt 2,\dots,x_{d-1} x_d \sqrt 2]^\top. 
\label{equ:phi2}
\end{equation}
Similarly, the simplified exact monomial embedding associated with $p=3$ is the function $\phib_3:\real^d \rightarrow \real^{(d^3+3 d^2+2 d)/{6}}$ defined as 
\begin{equation}
\phib_3(\xb) = [x_1^3,\dots,x_d^3,x_1^2x_2\sqrt 3,\dots,x_d^2 x_{d-1}\sqrt 3,x_1 x_2 x_3 \sqrt 6,\dots,x_{d-2} x_{d-1} x_d \sqrt 6]^\top. 
\label{equ:phi3}
\end{equation}


\section{Covariant aggregation of local descriptors}
\label{sec:method}

\begin{figure}[t]
\centering
\includegraphics[width=0.32\columnwidth]{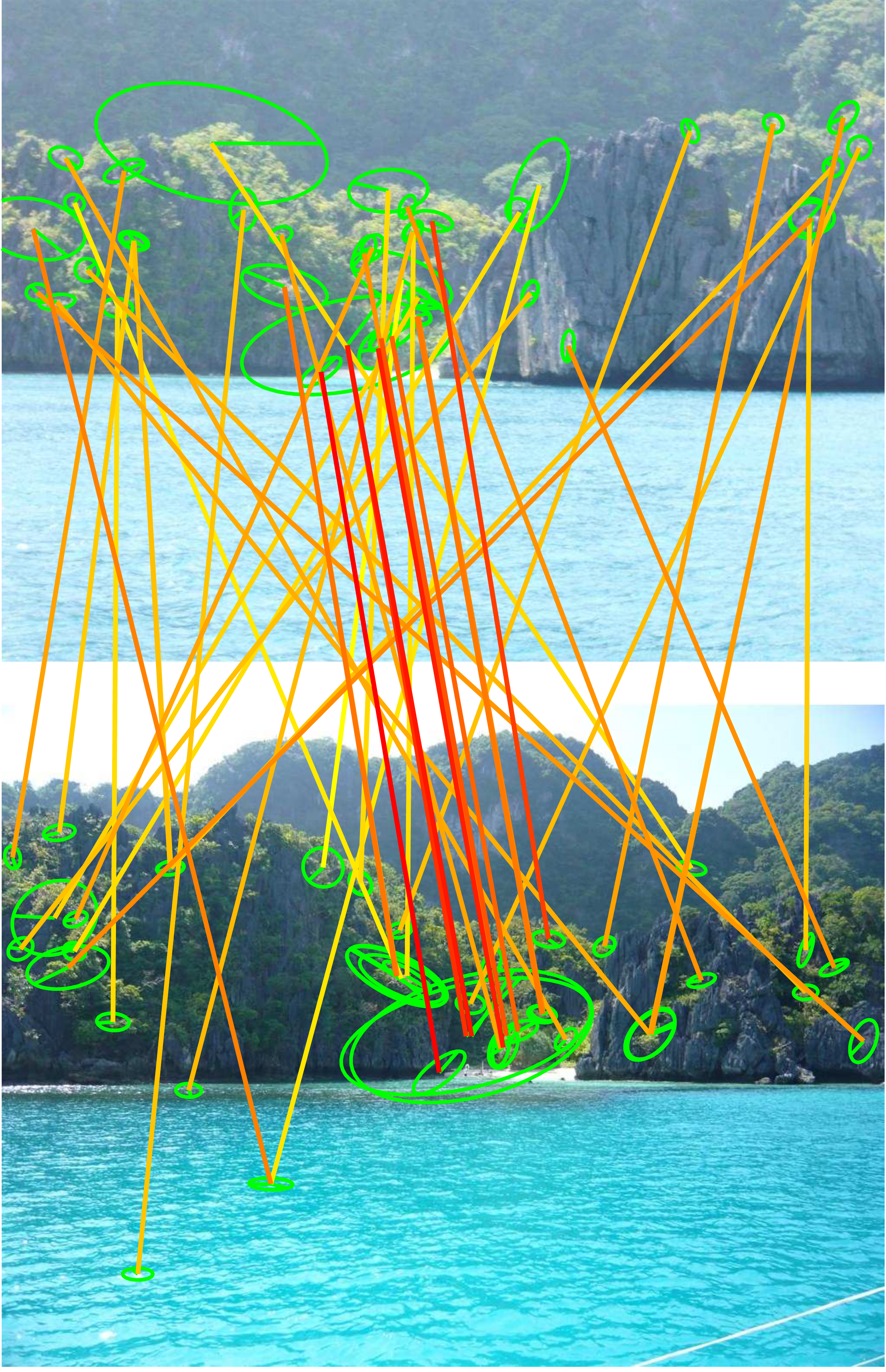} \hfill
\includegraphics[width=0.32\columnwidth]{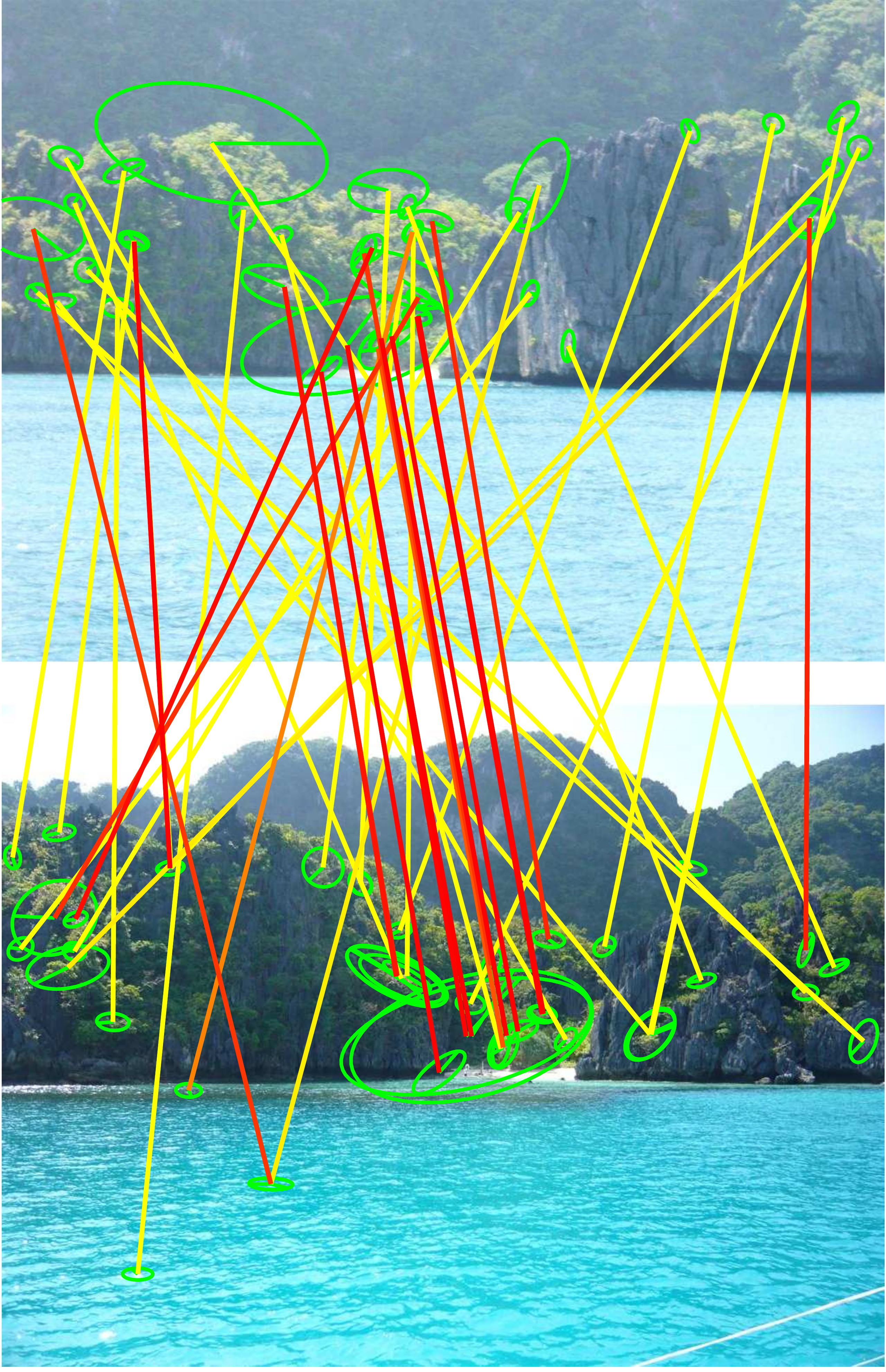} \hfill
\includegraphics[width=0.32\columnwidth]{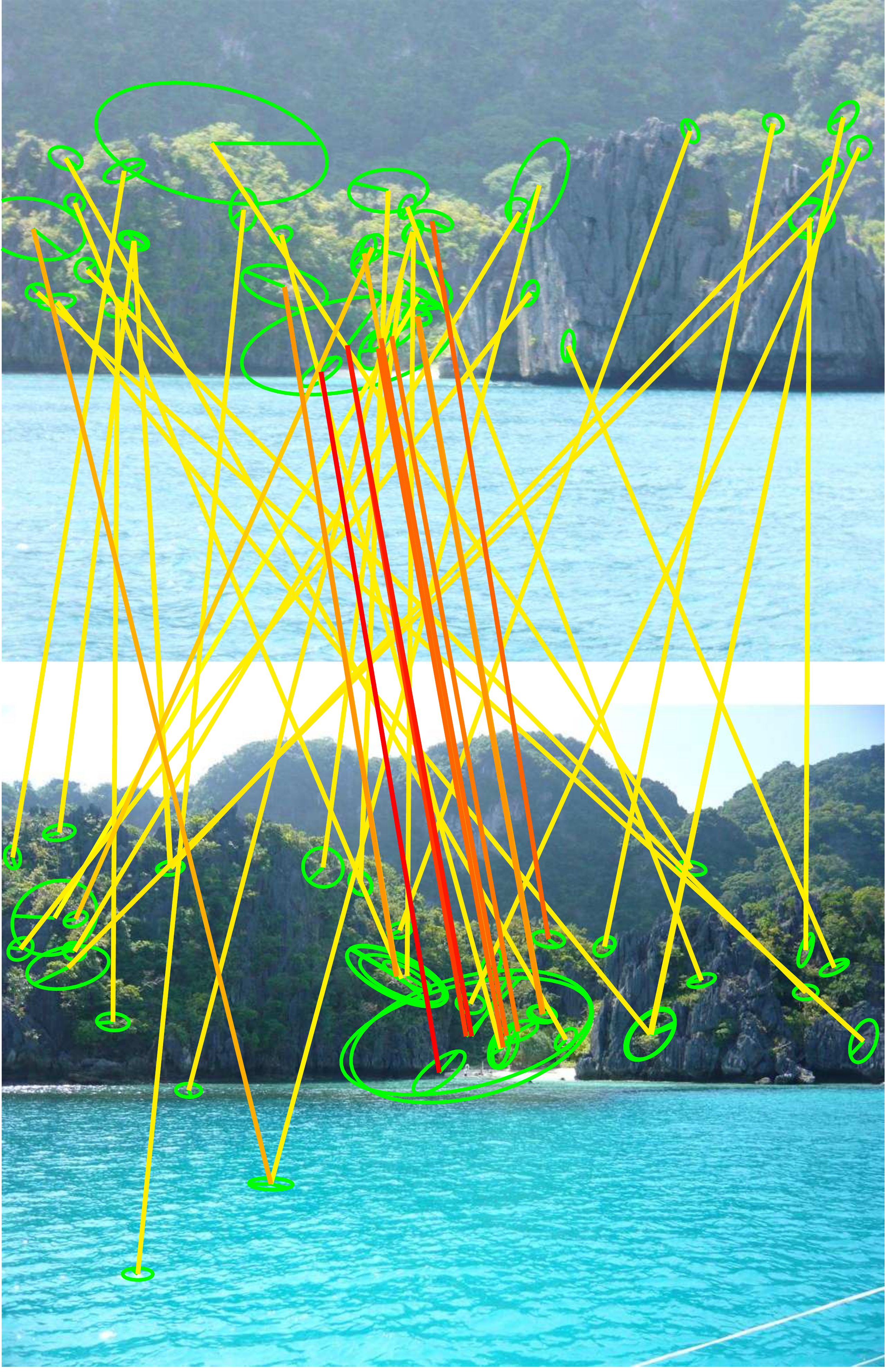}
\caption{Similarities between regions of interest, based on SIFT kernel~$k$ (\emph{left}),  angle consistency kernel $k_\theta$ (\emph{middle}) and both (\emph{right}). For each local region, we visualize the values $k(\xb,\yb)$, $k_\theta(\Delta\theta)$ and their product by the colors of the link (red=1).
\label{fig:matchexample}}
\end{figure}

The core idea of the proposed method is to exploit jointly the SIFT descriptors and the dominant orientation $\theta_x$ associated with a region of interest. For this purpose, we now assume that an image is represented by a set $\Xf$ of tuples, each of the form $(\xb,\theta_x)$, where $\xb$ is a SIFT descriptor and $\theta_x \in [-\pi,\pi]$ is the dominant orientation. 
Our objective is to obtain an approximation of a match kernel of the form 
\begin{align}
\Kf(\Xf,\Yf) & = \beta(\Xf) \beta(\Yf)
\sum_{(\xb,\theta_x) \in \Xf} \sum_{(\yb,\theta_y) \in \Yf} k(\xb,\yb) \, k_\theta(\theta_x,\theta_y) 
\label{equ:matchkernelangle} \\
 & = \langle\Xbf|\Ybf\rangle,
\label{equ:matchkernelangle2}
\end{align}
where $k$ is a local kernel identical to that considered in Section~\ref{sec:related} and $k_\theta$ reflects the similarity between angles. The interest of enriching this match kernel with orientation is illustrated by Figure~\ref{fig:matchexample}, where we show that several incorrect matches are downweighted thanks to this information. 

The kernel in~(\ref{equ:matchkernelangle}) resembles that implemented in WGC~\cite{JDS10a} with a voting approach. In contrast, we intend to approximate this kernel with an inner product between two vectors as in (\ref{equ:matchkernelangle2}), similar to the linear match kernel simplification in (\ref{equ:matchvector}). Our work is inspired by the kernel descriptors~\cite{BRF10} of Bo \etal, who also consider a kernel of a similar form, but at the patch level, to construct a local descriptor from pixel attributes, such as gradient and position. 

In our case, we consider the coding/pooling stage and employ a better approximation technique, namely explicit feature maps~\cite{VZ12}, to encode~$\Xf$. 
This section first explains the feature map of the angle, then how it modulates the descriptors, and finally discusses the match kernel design and properties. 

\subsection{A feature map for the angle}
The first step is to find a mapping $\alphab:[-\pi,\pi]\rightarrow\real^{M}$ from an angle $\theta$ to a vector $\alphab(\theta)$
such that $\alphab(\theta_{1})^{\top}\alphab(\theta_{2}) = k_\theta(\theta_{1}-\theta_{2})$. The function $k_\theta:\real\rightarrow[0,1]$ is a shift invariant kernel which should be symmetric ($k_\theta(\Delta\theta) = k_\theta(-\Delta\theta)$), pseudo-periodic 
with period of $2\pi$ and monotonically decreasing over $[0,\pi]$. 
We consider in particular the following function:
\begin{align}
k_{\VM}(\Delta\theta) = \frac{\exp(\kappa\cos(\Delta\theta))-\exp(-\kappa)}{2\sinh(\kappa)}.
\end{align}
It is derived from Von Mises distribution
$f(\Delta\theta;\kappa),$ 
which is often considered as the probability density distribution of the noise of the measure of an angle, and therefore regarded as the equivalent Gaussian distribution for angles. Although this is not explicitly stated in their paper, the regular Von Mises distribution is the kernel function implicitly used by Bo \etal~\cite{BRF10} for kernelizing angles. Our function $k_{\VM}$ is a shifted and scaled variant of Von Mises, designed such that its range is $[0,1]$, which ensures that $k_\VM(\pi)=0$. 

The periodic function $k_{\VM}$ can be expressed as a Fourier series whose coefficients
are (see~\cite{Abramowitz1964:Handbook}[Eq.~(9.6.19)]):
\begin{equation}
k_{\VM}(\Delta\theta) = \left(I_{0}(\kappa)-e^{-\kappa} + 2\sum_{n=1}^{\infty}I_{n}(\kappa)\cos(n\Delta\theta)\right).\frac{1}{2\sinh(\kappa)},
\end{equation}
where $I_{n}(\kappa)$ is the modified Bessel function of the first kind of order $n$. We now consider the truncation $\bar{k}^{N}_{\VM}$ of the series to the first $N$ terms:
\begin{equation}
\bar{k}^{N}_{\VM}(\Delta\theta) = \sum_{n=0}^{N}\gamma_{n} \cos(n\Delta\theta)
\text{\quad with }
\gamma_{0}=\frac{(I_{0}(\kappa)-e^{-\kappa})}{2\sinh(\kappa)} \text{ and }\gamma_{n}=\frac{I_{n}(\kappa)}{\sinh(\kappa)}\text{ if }n>0.
\end{equation}

We design the feature map $\alphab(\theta)$ as follows:
\begin{equation}
\label{eq:FeatureMap}
\alphab(\theta) = (\sqrt{\gamma_{0}}, \sqrt{\gamma_{1}}\cos(\theta),\ldots,\sqrt{\gamma_{N}}\cos(N\theta),\sqrt{\gamma_{1}}\sin(\theta),\ldots,\sqrt{\gamma_{N}}\sin(N\theta))^{\top}.
\end{equation}
This vector has $2N+1$ components. Moreover
\begin{align}
\alphab(\theta_{1})^{\top}\alphab(\theta_{2})&=\gamma_{0} + \sum_{n=1}^{N}\gamma_{n}(\cos(n\theta_{1})\cos(n\theta_{2})+\sin(n\theta_{1})\sin(n\theta_{2}))\\
&=\sum_{n=0}^{N} \gamma_{n}\cos(n(\theta_{1}-\theta_{2}))\\
&= \bar{k}^{N}_{\VM}(\theta_{1}-\theta_{2})\approx k_{\VM}(\theta_{1}-\theta_{2})
\end{align}
This process of designing a feature map is explained in full details by Vedaldi and Zisserman~\cite{VZ12}. This feature map gives an approximation of the target function $k_{\VM}$, which is more accurate as $N$ is bigger. 

\begin{figure}[t]
\begin{center}
\hfill
\includegraphics[width=0.32\columnwidth]{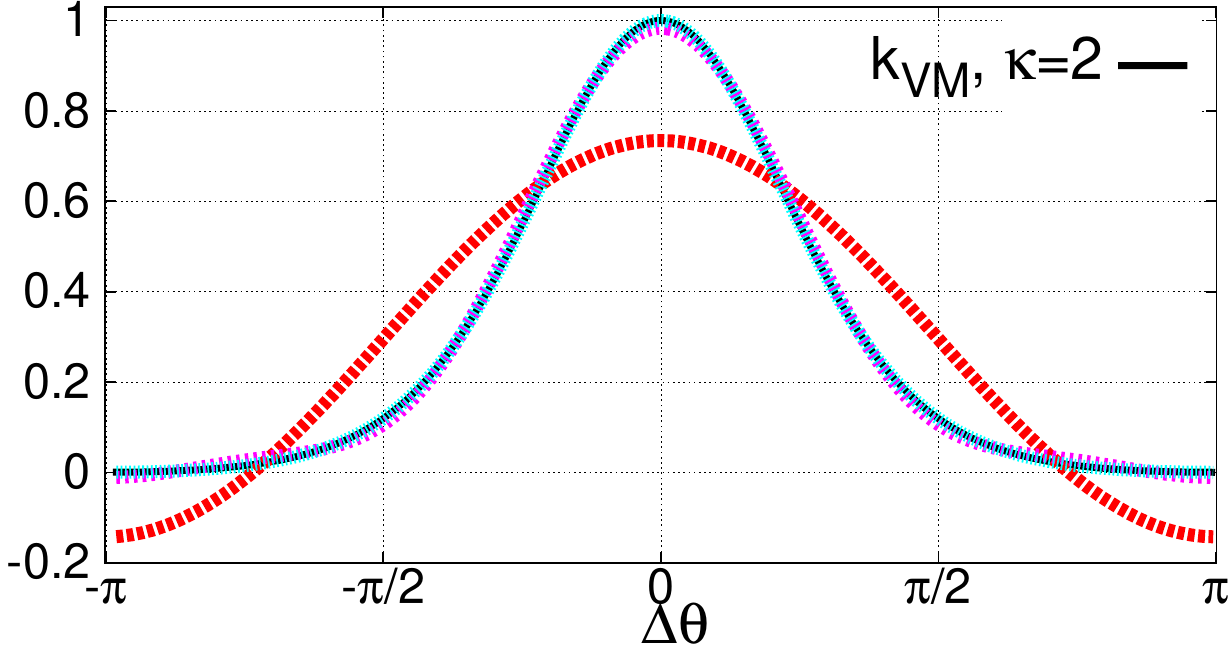}
\hfill
\includegraphics[width=0.32\columnwidth]{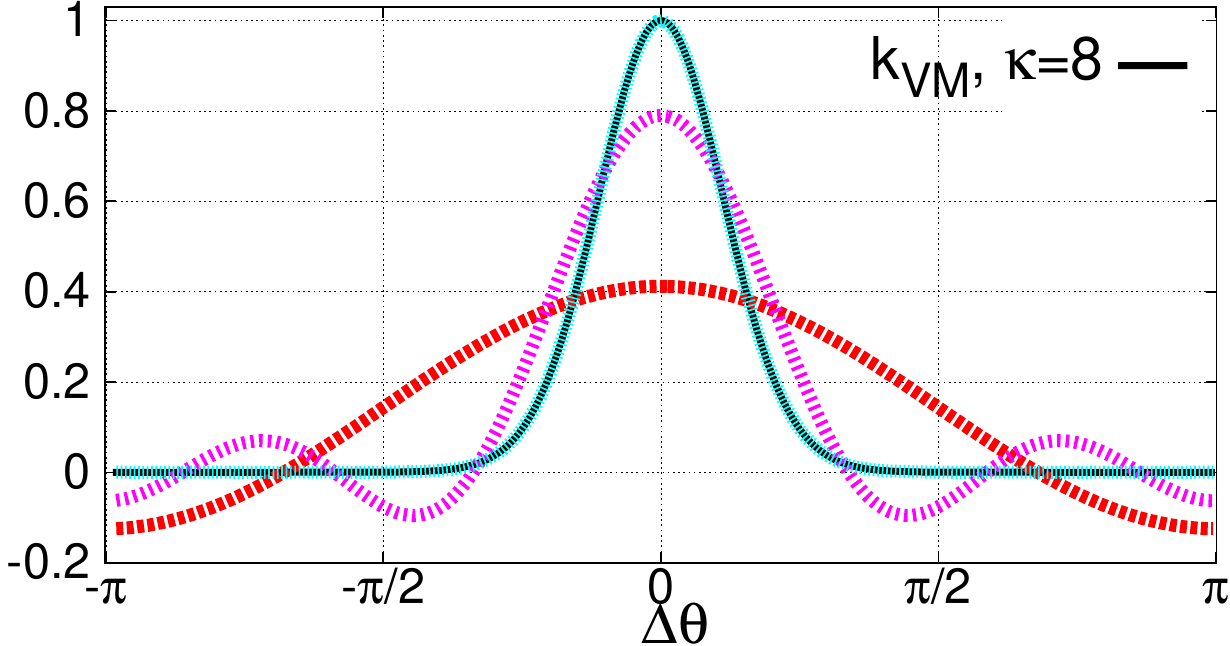}
\hfill
\includegraphics[width=0.32\columnwidth]{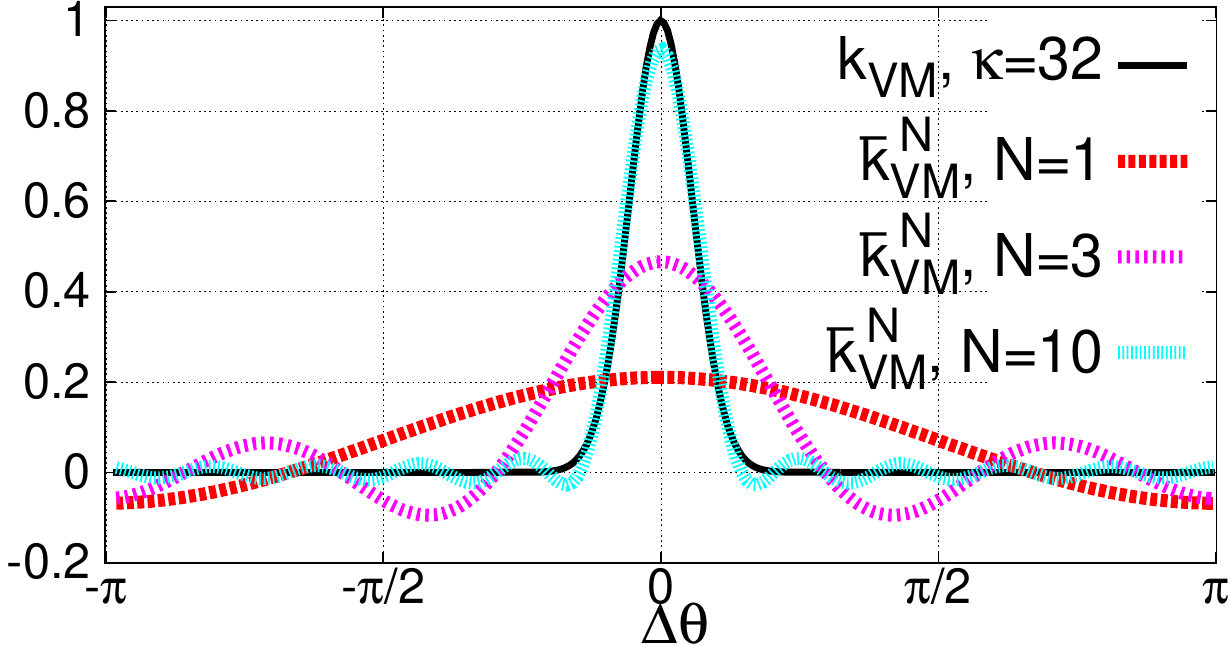}
\hfill
\end{center}
\vspace{-5ex}
\caption{Function $k_\VM$ for different values of $\kappa$ and its approximation $\bar{k}^{N}_{\VM}$ 
using $1$, $3$ and $10$ frequencies, as implicitly defined by the corresponding mapping $\alphab:[\pi,\pi] \rightarrow \real^{2N+1}$. 
\label{fig:angfun}}
\vspace{-2ex}
\end{figure}

Figure~\ref{fig:angfun} illustrates the function $k_\VM$ for several values of the parameter~$\kappa$ and its approximation $\bar{k}^{N}_{\VM}$  for different values of $N$. 
First note that $\bar{k}^{N}_{\VM}$ may not fulfill the original requirements:
its range might be wider than $[0,1]$ and it might not be monotonically decreasing over $[0,\pi]$. 
Larger values of $\kappa$ produce a more ``selective'' function of the angle, yet require more components (larger $N$) to obtain an accurate estimation. Importantly, the approximation stemming from this explicit angle mapping is better than that based on efficient match kernels~\cite{BS09}, which converges slowly with the number of components. Efficient match kernels are more intended to approximate kernels on vectors than on scalar values.  
As a trade-off between selectivity and the number of components, we set $\kappa$=8 and $N$=3 (see Section~\ref{sec:experiments}). Accordingly, we use $\bar{k}^{3}_{\VM}$ as $k_\theta$ in the sequel. The corresponding  embedding $\alphab:\real \rightarrow \real^7$ maps any angle to a 7-dimensional vector.

\paragraph{Remark:} Instead of approximating a kernel on angles with finite Fourier series, one may rather consider directly designing a function satisfying our initial requirements (pseudo-period, symmetric, decreasing over $[0,\pi]$), such as 
\begin{equation}
k_{P}(\Delta\theta) = \cos(\Delta\theta/2)^{P}\mbox{ with $P$ even}.
\end{equation}
This function, thanks to power reduction trigonometric identities for even $P$, is re-written as 
\begin{equation}
k_{P}(\Delta\theta) = \sum_{p=0}^{P/2}\gamma_{p} \cos(p\Delta\theta)
\end{equation}
\begin{equation}
\text{with \quad \quad \quad} \gamma_{0} = \frac{1}{2^{P}}{P\choose P/2}, \gamma_{p}=\frac{1}{2^{P-1}}{P\choose P/2-p}\quad0<p\leq P/2.
\end{equation}
Applying~\eqref{eq:FeatureMap} leads to a feature map $\alphab(\theta)$ with $P+1$ components such that
$\alphab(\theta_{1})^{\top}\alphab(\theta_{2}) = k_{P}(\theta_{1}-\theta_{2}).$
For this function, the interesting property is that the scalar product is exactly equal to the target kernel value $k_{P}(\theta_{1}-\theta_{2})$, and that the original requirements now hold. From our experiments, this function gives reasonable results, but requires more components than $\bar{k}_\VM$ to achieve a shape narrow around $\Delta\theta=0$ and close to 0 otherwise. The results for our image search application task using this function are slightly below our Von Mises variant for a given dimensionality. So, despite its theoretical interest we do not use it in our experiments. Ultimately, one would rather directly learn a Fourier embedding for the targeted task, in the spirit of recent works on Fourier kernel learning~\cite{BFS12}.

\subsection{Modulation and covariant match kernel}
The vector $\alphab$ encoding the angle $\theta$ ``modulates''\footnote{By analogy to communications, where modulation refers to the process of encoding information over periodic waveforms.}
any vector $\xb$ (or pre-mapped descriptor $\phib(\xb)$) with a function 
$m:\real^{2N+1}\times\real^{D}\rightarrow\real^{(2N+1) D}$. 
Thanks to classical properties of the Kronecker product $\otimes$, we have
\begin{equation}
m(\xb,\alphab(\theta)) = \xb\otimes\alphab(\theta) = (x_{1}\alphab(\theta)^{\top},x_{2}\alphab(\theta)^{\top}. \ldots,x_{d}\alphab(\theta)^{\top})^{\top}.
\end{equation}

We now consider two pairs of vectors and angle, $(\xb,\theta_{x})$ and $(\yb,\theta_{y})$, and their modulated descriptors 
$m(\xb,\alphab(\theta_{x}))$ and $m(\yb,\alphab(\theta_{y}))$. In the product space $\mathbb{R}^{(2N+1)D}$, the following holds:
\begin{align}
m(\xb,\alphab(\theta_{x}))^{\top}m(\yb,\alphab(\theta_{y})&=(\xb\otimes\alphab(\theta_{x}))^{\top}(\yb\otimes\alphab(\theta_{y}))\nonumber\\
& = (\xb^{\top}\otimes\alphab(\theta_{x})^{\top})(\yb\otimes\alphab(\theta_{y}))=(\xb^{\top}\yb)\otimes(\alphab(\theta_{x})^{\top}\alphab(\theta_{y}))\nonumber\\
& = (\xb^{\top}\yb)k_\theta(\theta_{x}-\theta_{y}).
\end{align}

\begin{figure}[t]
\begin{center}
\begin{tabular}{@{\nssp}c@{\nssp}@{\nssp}c@{\nssp}@{\nssp}c@{\nssp}@{\nssp}c@{\nssp}@{\nssp}c@{\nssp}@{\nssp}c@{\nssp}@{\nssp}c@{\nssp}@{\nssp}c@{\nssp}}
\includegraphics[width=0.13\columnwidth]{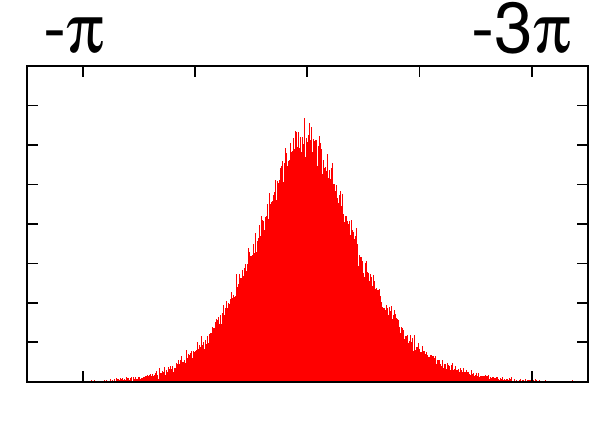}&
\includegraphics[width=0.13\columnwidth]{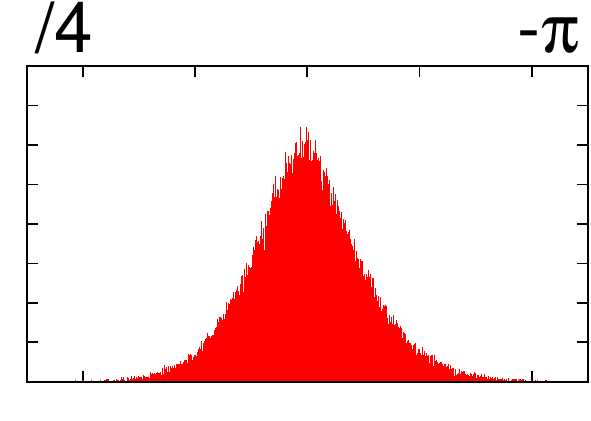}&
\includegraphics[width=0.13\columnwidth]{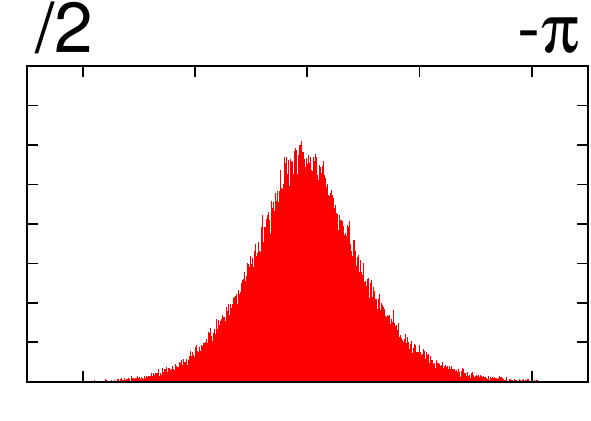}&
\includegraphics[width=0.13\columnwidth]{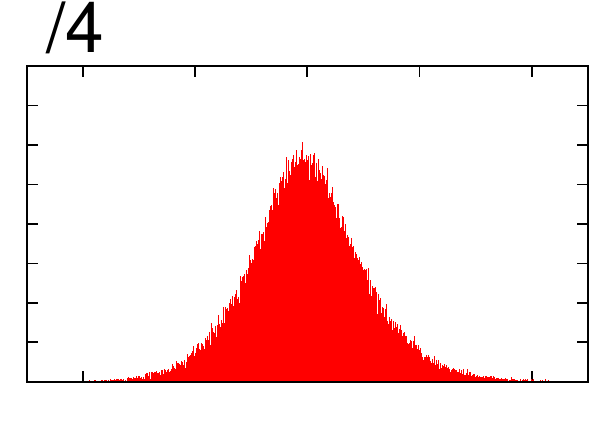}&
\includegraphics[width=0.13\columnwidth]{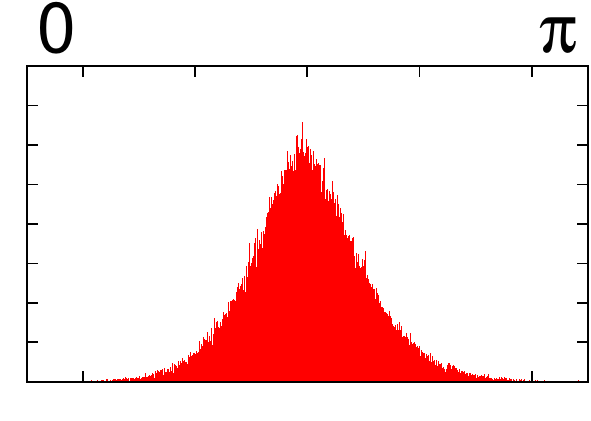}&
\includegraphics[width=0.13\columnwidth]{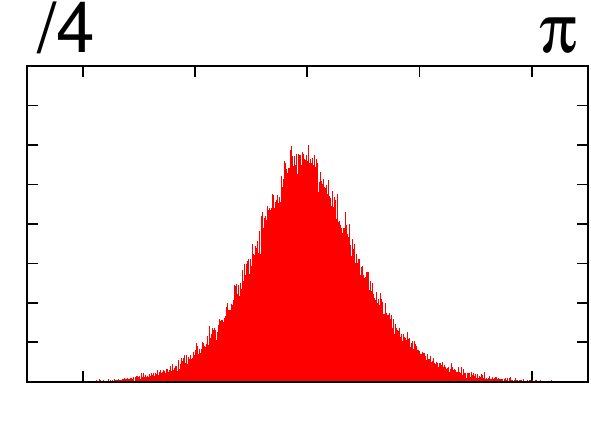}&
\includegraphics[width=0.13\columnwidth]{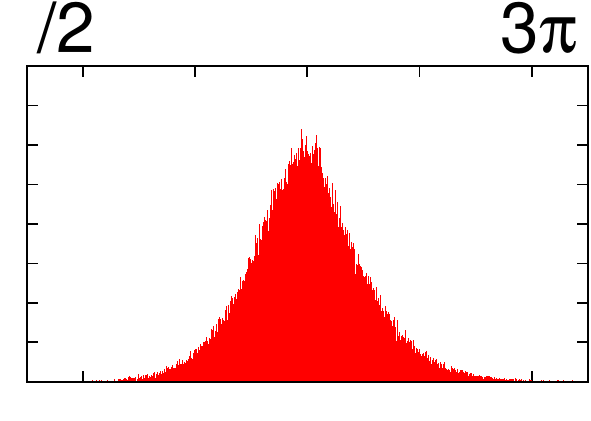}&
\includegraphics[width=0.13\columnwidth]{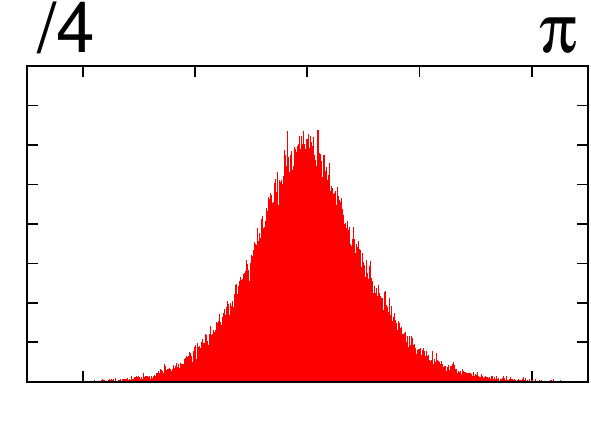}\\
\includegraphics[width=0.13\columnwidth]{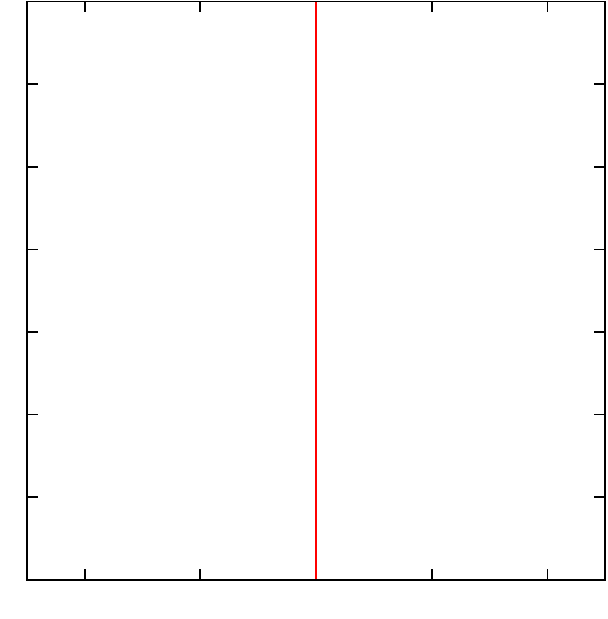}&
\includegraphics[width=0.13\columnwidth]{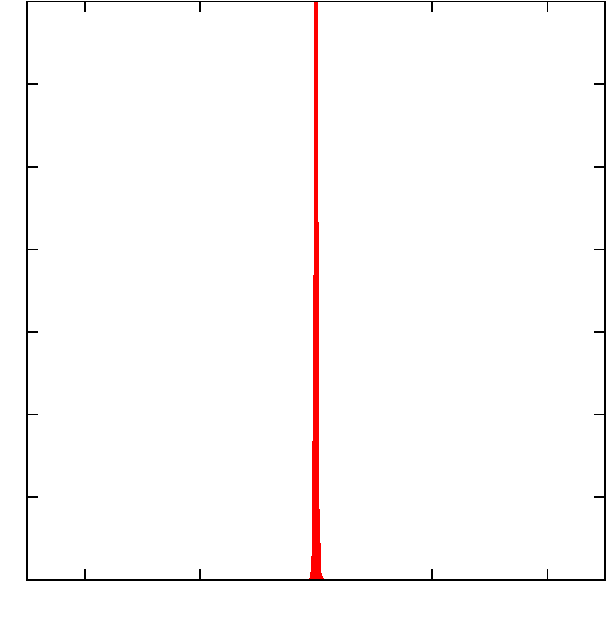}&
\includegraphics[width=0.13\columnwidth]{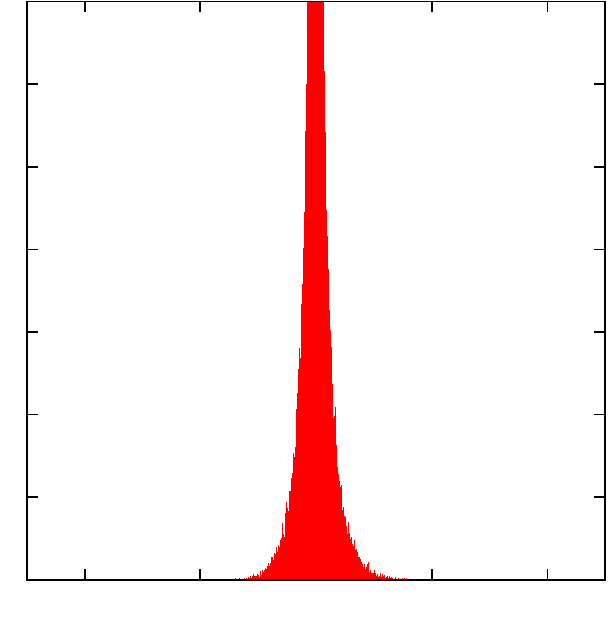}&
\includegraphics[width=0.13\columnwidth]{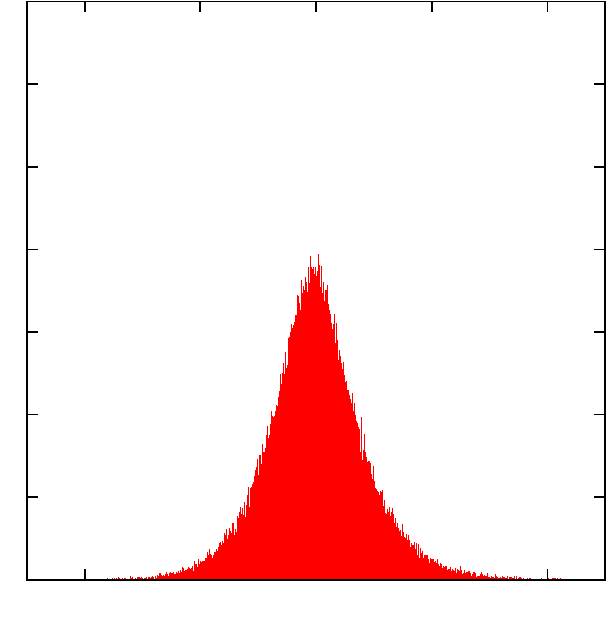}&
\includegraphics[width=0.13\columnwidth]{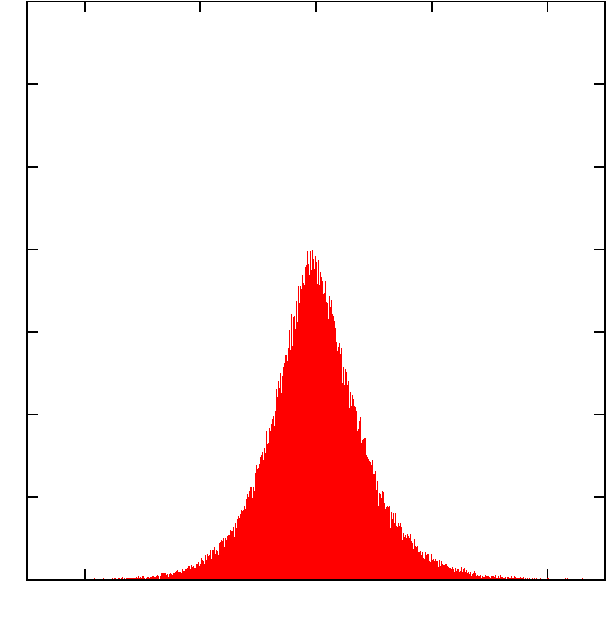}&
\includegraphics[width=0.13\columnwidth]{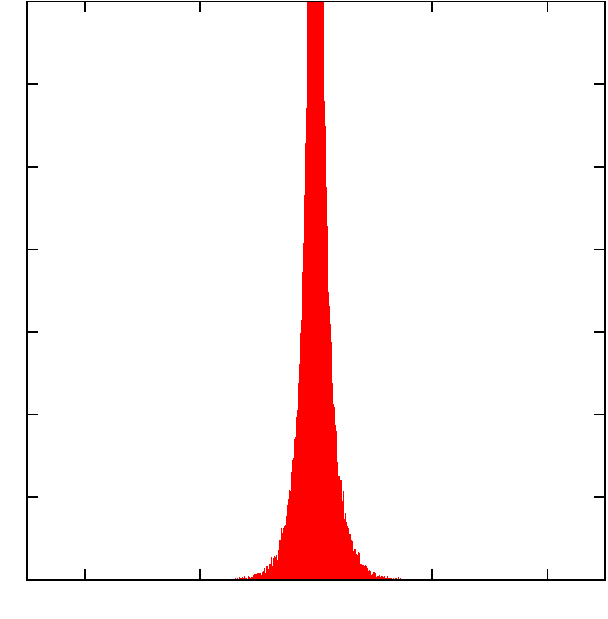}&
\includegraphics[width=0.13\columnwidth]{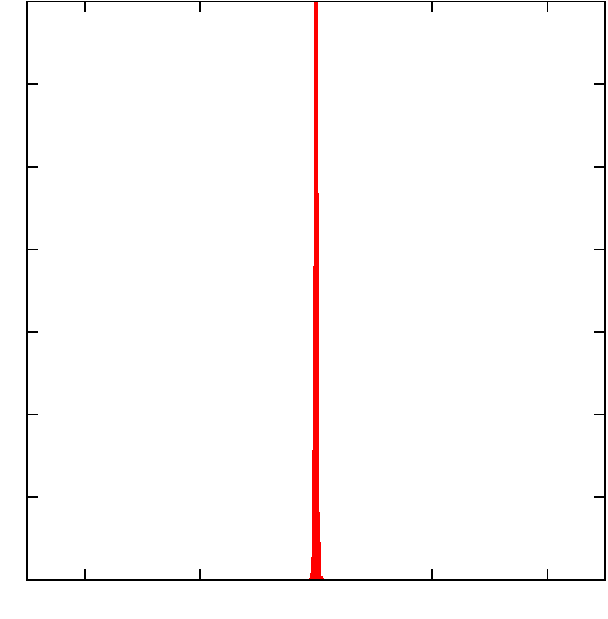}&
\includegraphics[width=0.13\columnwidth]{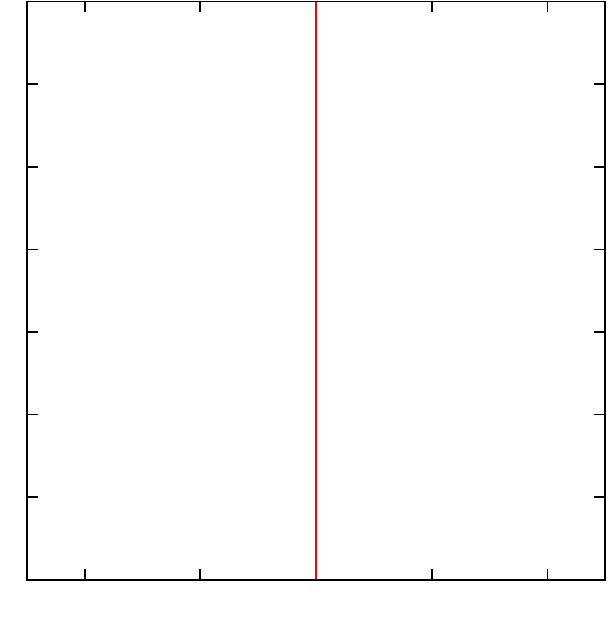}\\
\end{tabular}
\end{center}
\vspace{-5ex}
\caption{Distribution of patch similarity
for different values of orientation difference. In this figure, we split the angular space into $8$ equally-sized bins and present the similarity distribution separately for each of these bins. Horizontal axis represents the similarity value between matching features.\emph{Top}: distribution of similarities with kernel on SIFTs. \emph{Bottom:} Distribution after modulation with $\alpha$.
\label{fig:rotation_stats}}
\end{figure}

Figure~\ref{fig:rotation_stats} shows the distribution of the similarities between regions of interest before and after modulation, as a function of the difference of angles. Interestingly, there is no obvious correlation between the difference of angle and the SIFT: the similarity distribution based on SIFT is similar for all angles. This suggests that the modulation with angle provides complementary information.  

\paragraph{Combination with coding/pooling techniques.} Consider any coding method $\phib$ that can be written as match kernel (Fisher, LLC, Bag-of-words, VLAD, etc). The match kernel in~(\ref{equ:matchkernelangle}), with our $k_\theta$ approximation, is re-written as 
\begin{align}
\Kf(\Xf,\Yf) = & \beta(\Xf) \beta(\Yf)
\sum_{(\xb,\theta_{x}) \in \Xf} \sum_{(\yb,\theta_{y}) \in \Yf} m(\phib(\xb),\alphab(\theta_{x}))^{\top}m(\phib(\yb),\alphab(\theta_{y})), \nonumber \\
= & {\beta(\Xf) \left(\sum_{(\xb,\theta_{x})} m(\phib(\xb),\alphab(\theta_{x}))\right)^{\top}}
{\beta(\Yf) \left(\sum_{(\yb,\theta_{y})} m(\phib(\yb),\alphab(\theta_{y})\right)},
\label{equ:matchkernelvector}
\end{align}
where we observe that the image can be represented as the summation $\Xbf$ of the embedded descriptors modulated by their corresponding dominant orientation, as
\begin{equation}
\Xbf=\beta(\Xf) \sum_{(\xb,\theta_{x}) \in \Xf} m(\phib(\xb),\alphab(\theta_{x})). 
\end{equation}
This representation encodes the relative angles and is already more discriminative than an aggregation that does not consider them. However, at this stage, the comparison assumes that the images have the same global orientation. This is the case on benchmarks like Oxford5k building, where all images are orientated upright, but this is not true in general for particular object recognition. 

\subsection{Rotation invariance} 
\label{sec:rotationinvariance}
We now describe how to produce a similarity score when the orientations of related images may be different.
We represent the image vector $\Xbf$ as the concatenation of $2N+1$ $D$-dimensional subvectors associated to one term of the finite Fourier series: $\Xbf=[{\Xbf_0}^\top,{\Xbf_{1,c}}^\top,{\Xbf_{1,s}}^\top,\dots,{\Xbf_{N,c}}^\top,{\Xbf_{N,s}}^\top]^\top$. The vector $\Xbf_0$ is associated with the constant term in the Fourier expansion, $\Xbf_{n,c}$ and $\Xbf_{n,s}$, $1\leq n \leq N$, correspond to the cosine  and sine terms, respectively. 

Imagine now that this image undergoes a global rotation of angle $\theta$. Denote $\Xsetrot$ the new set of pairs $(\xb,\breve{\theta}_x)$ with
$\breve{\theta}_x=\theta_x-\theta$, and $\Xrot$ is the new image vector derived from these local descriptors. It occurs that ${\Xrot_0}={\Xbf_0}$ because this term does not depend on the angle, and that, for a given frequency bin $n$, elementary trigonometry identities lead to
\begin{align}
\Xrot_{n,c} & = \Xbf_{n,c} \cos n\theta + \Xbf_{n,s} \sin n\theta \\
\Xrot_{n,s} & = -\Xbf_{n,c} \sin n\theta + \Xbf_{n,s} \cos n\theta.
\end{align}
This in turn shows that  $\|\Xrot\| = \|\Xbf\|$. Therefore the rotation has no effect on the normalization factor $\beta(\Xf)$.

When comparing two images with such vectors, the linearity of the inner product ensures that  
\begin{align}
\langle \Xrot | \Ybf \rangle = \langle \Xbf_0 | \Ybf_0 \rangle 
& +  \sum_{n=1}^N \cos n\theta\ \left( 
  \langle \Xbf_{n,c} | \Ybf_{n,c} \rangle + \langle \Xbf_{n,s} | \Ybf_{n,s} \rangle \right) \\
& +  \sum_{n=1}^N \sin n\theta\ \left( - \langle \Xbf_{n,c} | \Ybf_{n,s} \rangle
  + \langle \Xbf_{n,s} | \Ybf_{n,c} \rangle \right).
\end{align}

Here, we stress that the similarity between two images is a real trigonometric polynomial in $\theta$ (rotation angle) of degree $N$. Its $2N+1$ components are fully determined by computing $\langle\Xbf_0|\Ybf_0\rangle$ and the inner products between the subvectors associated with each frequency, \ie, $\langle \Xbf_{n,c} | \Ybf_{n,c} \rangle$, $\langle \Xbf_{n,s}|\Ybf_{n,s}\rangle$, $\langle \Xbf_{n,c}|\Ybf_{n,s}\rangle$ and $\langle \Xbf_{n,s}|\Ybf_{n,c}\rangle$. Finding the maximum of this polynomial amounts to finding the rotation maximizing the score between the two images. 

Computing the coefficients of this polynomial requires a total of $D\times (1+4N)$ elementary operations for a vector representation of dimensionality $D\times (1+2N)$, that is, less than twice the cost of the inner product between $\Xbf$ and $\Ybf$.  
Once these components are obtained, the cost of finding the maximum value achieved by this polynomial is negligible for large values of $D$, for instance by simply sampling a few values of $\theta$. Therefore, if we want to offer the orientation invariant property, the complexity of similarity computation is typically twice the cost of that of a regular vector representation (whose complexity is equal to the number of dimensions). 

\paragraph{Remark:} This strategy for computing the scores for all possible orientations of the query is not directly compatible with non-linear post-processing of $\Xbf$ such as component-wise power-law normalization~\cite{PSM10}, except for the subvector $\Xbf_0$. We propose two possible options to overcome this problem. 
\begin{enumerate}
\item The naive strategy is to compute the query for several hypothesis of angle rotation, typically 8. In theory, this multiplies the query complexity by the same factor 8. However, in practice, it is faster to perform the matrix-matrix multiplication, with the right matrix representing 8 queries, than computing separately the corresponding 8 matrix-vector multiplications. 
We use this simpler approach in the experimental section. 
\item Alternately, the power-law normalization is adapted to become compatible with our strategy: we compute the modulus of the complex number represented by two components ($\sin$ and $\cos$) associated with the same frequency $n$ and the same original component in $\phib(\xb)$. These two components are then divided by the square-root (or any power) of this modulus. Experimentally, this strategy is as effective as the naive option. 
\end{enumerate}

\section{Experiments}
\label{sec:experiments}

We evaluate the performance of the proposed approaches and compare with state of the art methods 
on two publicly available datasets for image and particular object retrieval, namely Inria Holidays~\cite{JDS08} and Oxford Buildings 5k~\cite{PCISZ07}. 
We also combine the latter with 100k distractor images to measure the performance on a larger scale. The merged dataset is referred to as Oxford105k. The retrieval performance is measured with mean Average Precision (mAP)~\cite{PCISZ07}.

Our approach modulates any coding/pooling technique operating as a match kernel. Therefore, we evaluate the benefit of our approach combined with several coding techniques, namely
\begin{list}{$\circ$}{}
\item VLAD~\cite{JPDSPS11}, which encodes a SIFT descriptor by considering the residual vector to the centroid. 
\item The Fisher vector~\cite{PD07,PSM10,JH98}. For image classification, Chatfield \etal~\cite{CLVZ11} show that it outperforms concurrent coding techniques, in particular LLC~\cite{WYY10}. We adopt the standard choice for image retrieval and use only the gradient with respect to the mean~\cite{JPDSPS11}.
\item Monomomial embeddings of order 2 and 3 applied on local descriptors (See below for pre-processing), \ie, the functions $\phib_2$ in (\ref{equ:phi2}) and $\phib_3$ in (\ref{equ:phi3}). For the sake of consistency, we also denote by $\phib_1$ the function $\phib_1:x\rightarrow x$. 
\end{list}
We refer to these methods combined with our approach with the symbol~``$\otimes$'': 
VLAD$\otimes$, Fisher$\otimes$, $\phib_1\otimes$, $\phib_2\otimes$ and $\phib_3\otimes$, correspondingly. 
In addition, we compare against the most related work, namely the recent CVLAD~\cite{ZJG13} method, which also aims at producing an image vector representation integrating the dominant orientations of the patches. 
Whenever the prior work is not referenced, results are produced using our own (improved) implementations of VLAD, Fisher and CVLAD, so that the results are directly comparable with the same features.

\subsection{Implementation Details}

\paragraph{Local descriptors. }
We  use the Hessian-Affine detector~\cite{MTSZMSKG05} to extract the regions of interest, 
that are subsequently described by SIFT descriptors~\cite{L04} post-processed with 
RootSIFT~\cite{AZ12}. Then, following the pre-processing required for the Fisher 
vector~\cite{PD07,PSM10,JPDSPS11}, we apply PCA to reduce the vector to 80 components.
An exception is done for VLAD and CVLAD with which we only use the PCA basis to center and rotate
descriptors as suggested by Delhumeau~\cite{DGJP13}, without dimensionality reduction. 
The resulting vector is subsequently $\ell_2$-normalized. 

The improved Hessian-Affine detector of Perdoch~\etal~\cite{PCM09} improves the retrieval performance. However, we do not use it, since it ignores rotations by making the gravity vector assumption. Instead, we use the original detector modified so that it has similar parameters (patch size set to 41). 

\paragraph{Codebook.} For all methods based on codebooks, we only consider distinct datasets for learning. More precisely and following common practice, the k-means and GMM (for VLAD and Fisher, respectively) are learned on Flickr60k for Inria Holidays and Paris6k~\cite{PCISZ08} for Oxford buildings. We rely on the Yael library~\cite{DJ14} for codebook construction and VLAD and Fisher encoding.

\paragraph{Post-processing.}
The final image vector obtained by each method is power-law normalized~\cite{JDS09a,PSM10,JPDSPS11}. This processing improves the performance by efficiently handling the burstiness phenomenon.
Exploiting the dominant orientation in our covariant match kernel provides a complementary way to further handle the same problem. We mention that using the dominant orientation is shown effective in a recent work by Torii~\etal~\cite{TSPO13}. With our angle modulation, this post-processing inherently captures and down weights patches with similar dominant orientation. 
The power-law exponent is set to $0.4$ for Fisher and VLAD and to $0.2$ for monomial embeddings. These values give best or close-to-best performance for the initial representations. The resulting vector is  $\ell_2$-normalized. 

In addition to power-law normalization, we rotate the aggregated vector representation with a PCA rotation matrix~\cite{JC12,SQ13}. This aims at capturing the co-occurrences to down-weight them either by whitening~\cite{JC12} or a second power-law normalization~\cite{SQ13}. We adopt the latter choice (with exponent 0.5) to avoid the sensitivity to eigenvalues (in whitening) when learning PCA with few input data. 
We refer to this Rotation and Normalization as RN in our experiments.

Optionally, to produce compact representations, we keep only the first few components (the most energetic ones) and $\ell_2$-normalize the shortened vector. 

\paragraph{Query rotation.}
In order to obtain rotation invariance jointly with power-law normalization and RN, we apply rotations of the query image and apply individual queries as described in Section~\ref{sec:method} (option 1). We apply 8 query rotations on Holidays dataset. On Oxford5k, we rather adopt the common choice of not considering other possible orientations: Possible rotation of the query object is usually not considered since all the images are up-right.

\begin{figure}[t]
\includegraphics[width=0.42\columnwidth]{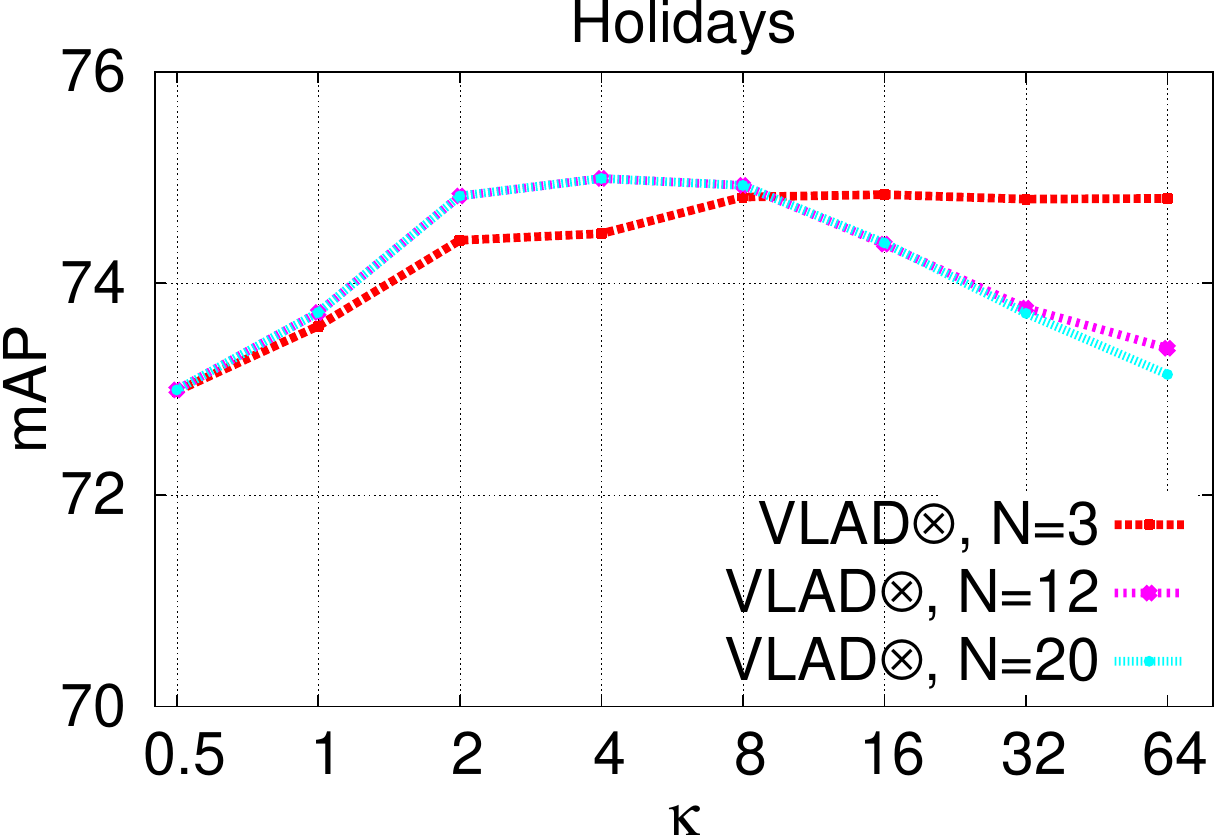} 
\hfill
\includegraphics[width=0.42\columnwidth]{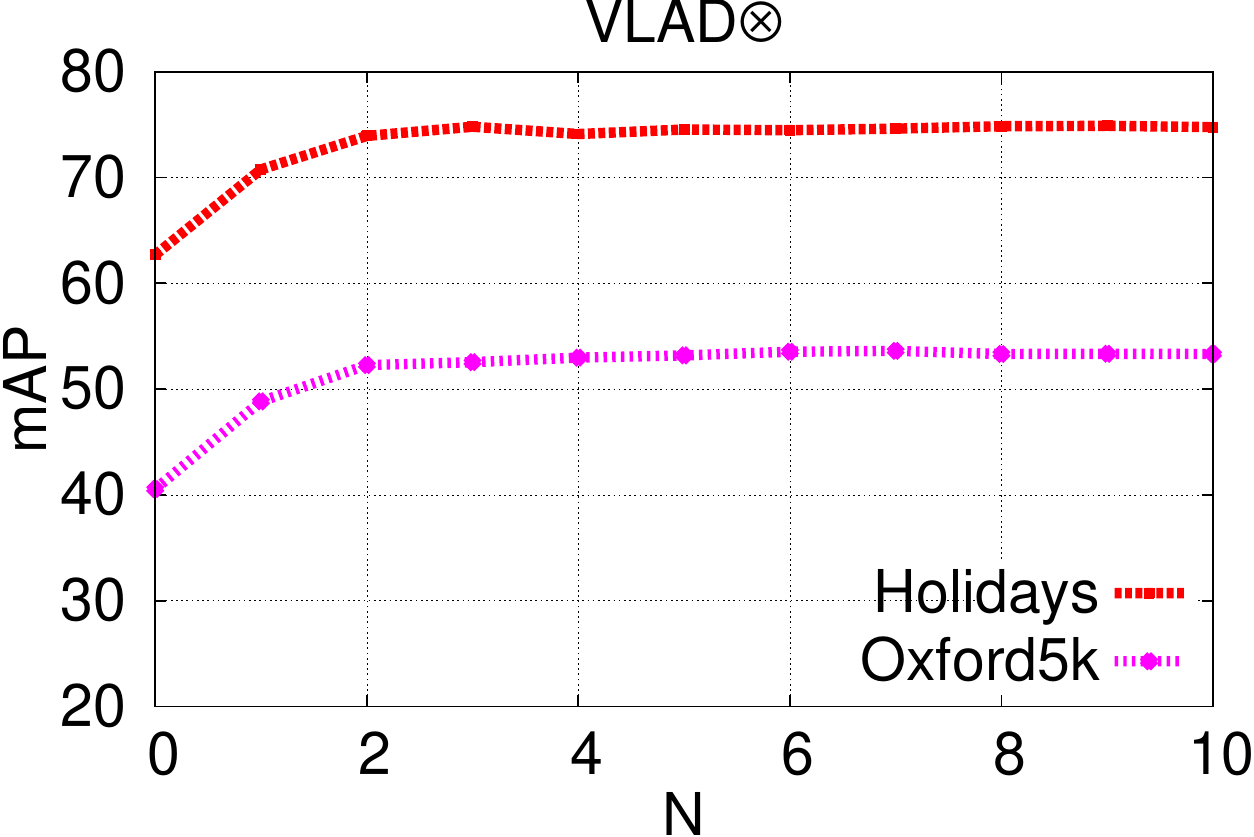}
\vspace{-2ex}
\caption{Left: Performance on Holidays dataset of modulated VLAD for different values of $\kappa$ and for different approximations. 
Right: Performance comparison of modulated VLAD for increasing number of components of the angle feature map. Zero corresponds to original VLAD (not modulated). A codebook of 32 visual words is used.}
\label{fig:compareafun}
\end{figure}

\subsection{Impact of the parameters}
The impact of the angle modulation is controlled by the function $k_\theta$ parametrized by $\kappa$ and $N$. 
As shown in Figure~\ref{fig:angfun}, the value $\kappa$ typically controls the ``bandwitdh'', \ie, the range of $\Delta\theta$ values with non-zero response. 
The parameter $N$ controls the quality of the approximation, and implicitly constrains the achievable bandwidth. It also determines the dimensionality of the output vector. 

Figure~\ref{fig:compareafun} (left) shows the impact of these parameters on the performance. 
As to be expected, there is a trade-off between defining too narrow or too large. The optimal performance is achieved with $\kappa$ in the range $[2,8]$. 
Figure~\ref{fig:compareafun} (right) shows the performance for increasing number of frequencies, which rapidly converges to a fixed mAP. 
This is the mAP of the exact evaluation of (\ref{equ:matchkernelangle}). We set $N=3$ as a compromise between dimensionality expansion and performance. Therefore the modulation multiplies the input dimensionality by 7. 

\subsection{Benefit of our approach}

\begin{table}[t]
\centering
\setlength\extrarowheight{-0.5pt}
{\small
\begin{tabular}{|@{\sssp}l@{\sssp}|r@{\sssp}|r@{\msp}r@{\msp}r@{\sssp}|r@{\msp}r@{\sssp}|r@{\msp}r@{\msp}r@{\msp}r@{\sssp}|r@{\sssp}|r@{\sssp}|}
\hline
Method        &               \multicolumn{1}{c|}{$\phib_1$}      & \multicolumn{3}{c|}{$\phib_1\otimes$}       & \multicolumn{2}{c|}{$\phib_2$} & \multicolumn{4}{c|}{$\phib_2\otimes$} & \multicolumn{1}{c|}{$\phib_3$} & \multicolumn{1}{c|}{$\phib_3\otimes$} \\         \hline
RN            &          &       &        &                &         &$\times$\sssp\, &         &       &$\times$\sssp\, & $\times$\sssp\, &  & \\
$N$           & --\sssp~ &   1   &     3  &  6             &  --\sssp~ & --\sssp~        &   1     &   3   &  1     & 3     & --\sssp~ & 1  \\
\#dim         &  80      &  240  &   560  & 1,040          &  3240   & 3,240  &  9,720   & 22,680 &  9,720  & 22,680  & 88,560 & 265,680 \\
\hline
mAP           &  35.4    &  48.9 &  59.5  & \textbf{63.2}  &  59.7   & 71.6   & 68.8    & 73.7  &  75.3  &  \textbf{79.9}  & 60.0 & 72.5 \\
\hline
\end{tabular}}
\smallskip
\vspace{-1ex}
\caption{Impact of modulation on monomial embeddings of order 1, 2 and 3.
The performance is reported for Holidays dataset.
RN = Rotation and Normalization. 
\label{tab:mono}}
\vspace{-3ex}
\end{table}

\begin{figure}[t]
\includegraphics[width=0.45\columnwidth]{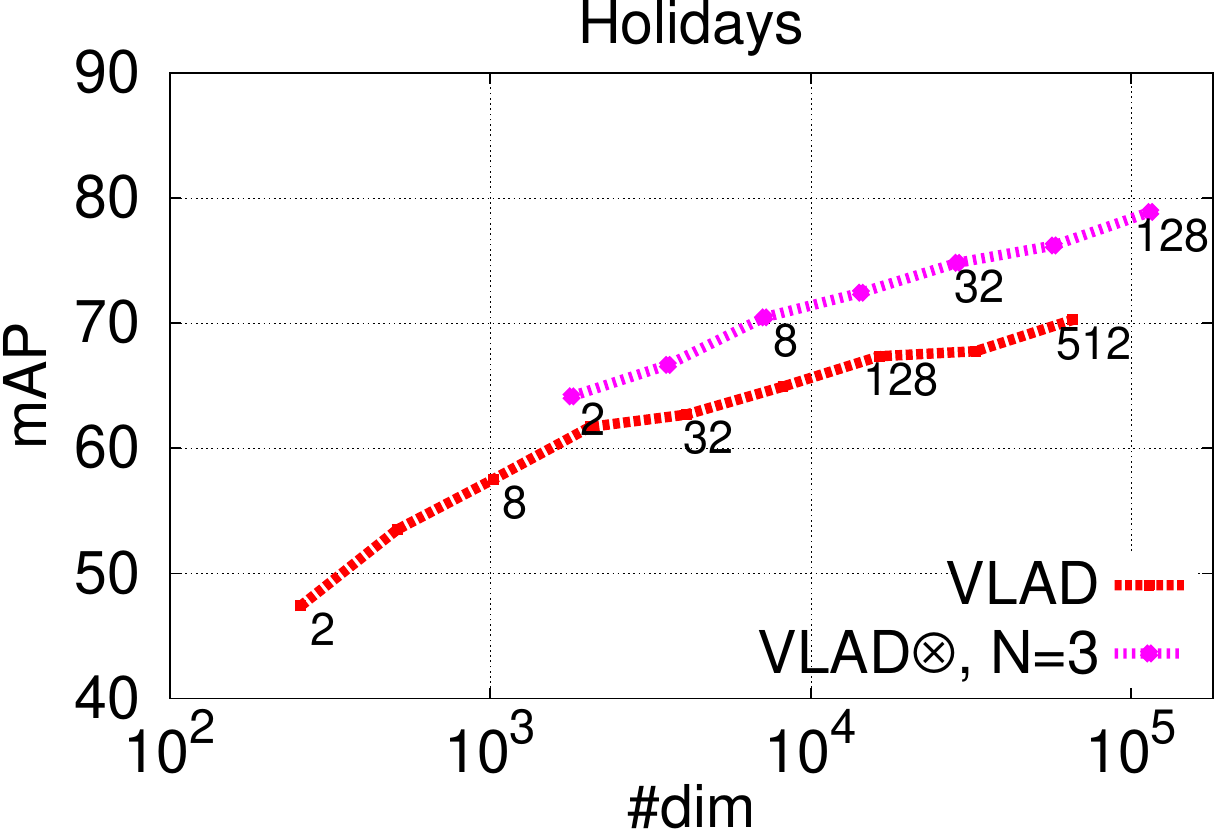}
\hfill
\includegraphics[width=0.45\columnwidth]{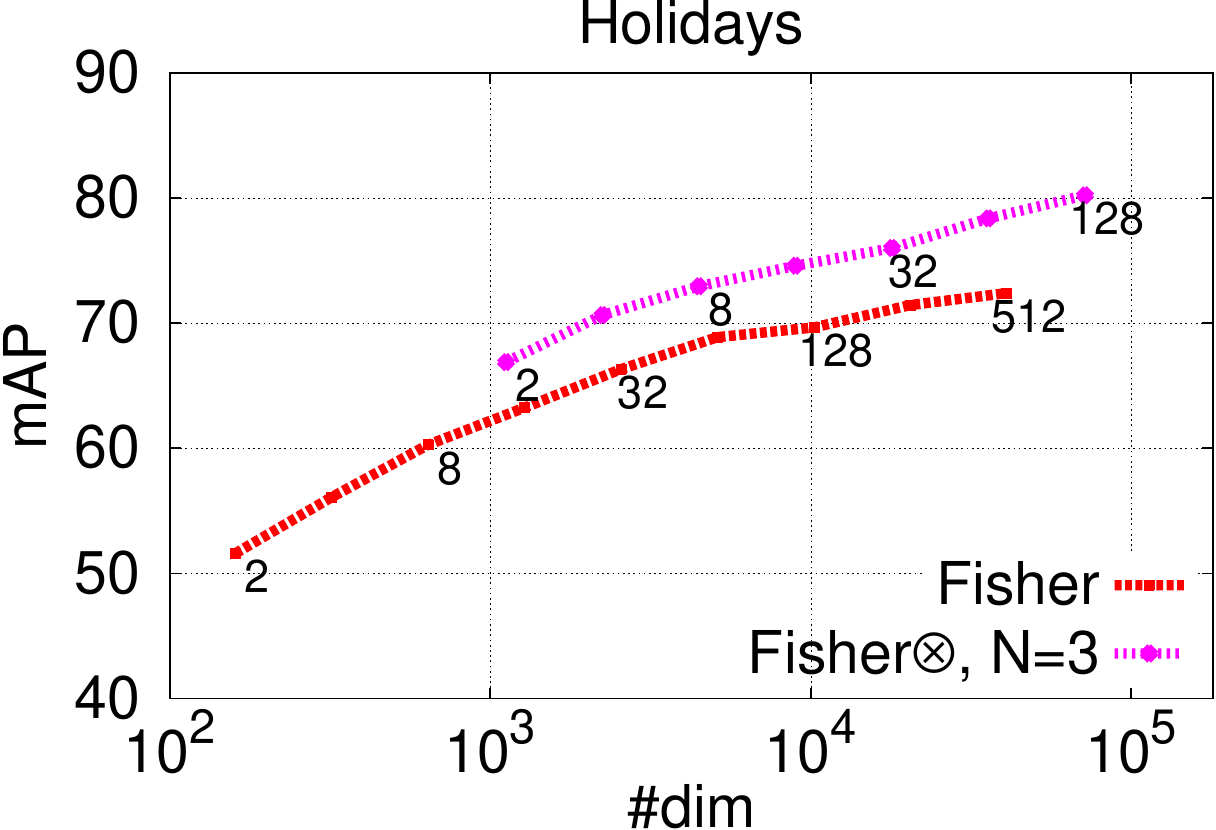}
\vspace{-2ex}
\caption{Impact of modulation on VLAD and Fisher: Performance versus dimensionality of the final
 vector for VLAD (left) and Fisher (right) compared to their 
modulated counterparts. Codebook size is shown with text labels. 
Results for Holidays dataset.\label{fig:impvladfisher}}\vspace{-2ex}
\end{figure}

Table~\ref{tab:mono} shows the benefit of modulation when applied to the monomial embeddings $\phib_1$, $\phib_2$ and $\phib_3$. The results are on par with the recent coding techniques like VLAD or Fisher improved with modulation. We consider the obtained performance as one of our main achievements, because the representation is codebook-free and requires no learning. 
In addition, we further show the benefit of combining monomial embeddings with RN. This significantly boosts performance with the same vector dimensionality and negligible computational overhead.

We compare VLAD, Fisher and monomial embeddings to their modulated counterparts.
Figure~\ref{fig:impvladfisher} shows that modulation significantly improves the performance for the same codebook size. However, given that the modulated vector is $\times 7$ larger (with $N=3$), the comparison focuses on the performance obtained with the same dimensionality. Even in this case, modulated VLAD$\otimes$ and Fisher$\otimes$ offer a significant improvement. We can conclude that it is better to increase the dimensionality by modulation than using a larger codebook.

\begin{figure}
\includegraphics[width=0.45\columnwidth]{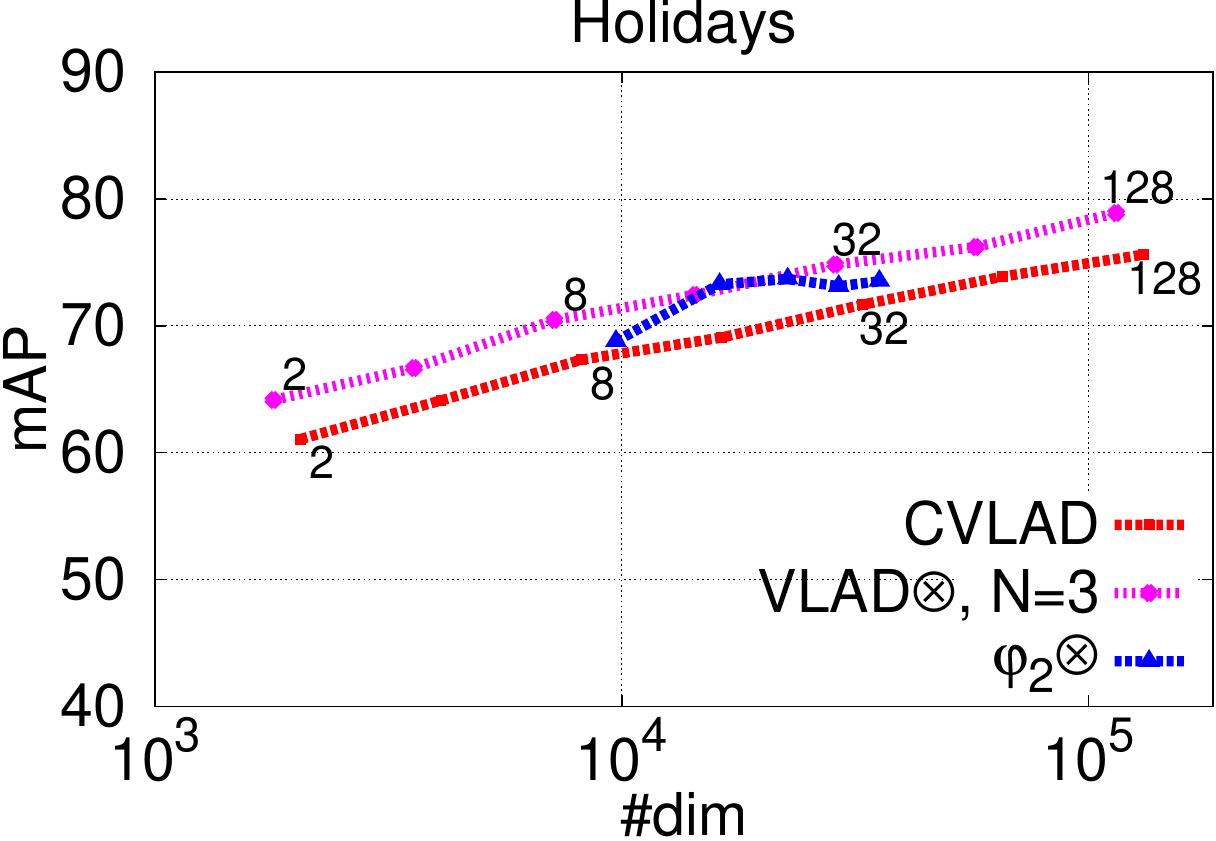}
\hfill
\includegraphics[width=0.45\columnwidth]{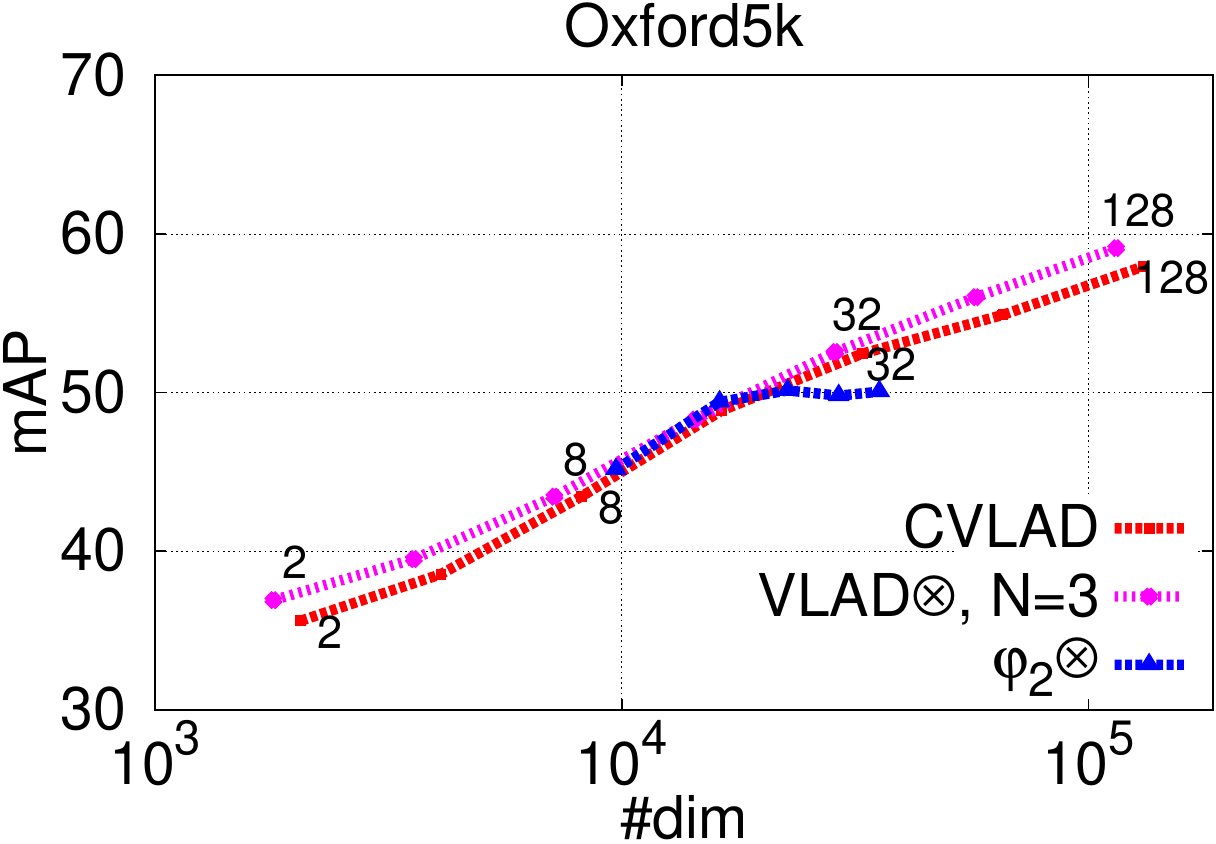}
\vspace{-2ex}
\caption{Comparison to CVLAD. We measure performance on Holidays and Oxford5k for CVLAD and our proposed methods for increasing codebook size. The codebook cardinality is shown with text labels for CVLAD and modulated VLAD, while for $\phib_2$ the  number of frequency components ($N$) used are increased from $1$ to $5$.\label{fig:comparecvlad}}
\end{figure}

\vspace{-3ex}

\subsection{Comparison to other methods}
\vspace{-1.5ex}

\begin{table}
\setlength\extrarowheight{-0.5pt}
\centering
\begin{tabular}{|@{\msp}l@{\msp}|@{\msp}c@{\msp}r@{\msp}|@{\msp}c@{\msp}|@{\msp}c@{\msp}c@{\msp}c@{\msp}|} 
\hline  
Method                    &  \#C  &\#dim  &   RN   &  Holidays    &  Oxford5k   & Oxford105k  \\ \hline \hline
VLAD~\cite{JPDSPS11}      &   64  & 4,096 &        & 55.6         & 37.8        &   -      \\  
Fisher~\cite{JPDSPS11}    &   64  & 4,096 &        & 59.5         & 41.8        &   -      \\  
VLAD~\cite{JPDSPS11}      & 256 & 16,384  &        & 58.7         & -     &   -      \\  
Fisher~\cite{JPDSPS11}    & 256 & 16,384  &        & 62.5         & -       &   -      \\  
Arandjelovic~\cite{AZ13}    & 256   & 32,536  &        & 65.3         &\textbf{55.8}&   -     \\  
Delhumeau~\cite{DGJP13}   & 64    & 8,192 &        & 65.8         & 51.7        &       45.6      \\ 
Zhao~\cite{ZJG13}           & 32    & 32,768  &        &\textbf{68.8} & 42.7        &       -     \\ \hline
VLAD$\otimes$             & 32    & 28,672  &        & 74.8         &  52.5       &      46.3\\   
VLAD$\otimes$             & 32    & 28,672  &$\times$& 81.0         &  61.8       &      \textbf{53.9}\\  
Fisher$\otimes$           & 32    & 17,920  &        & 76.0         &  51.0       &      44.9\\   
Fisher$\otimes$           & 32    & 17,920  &$\times$& 81.2         &  60.7       &      52.2\\ 
Fisher$\otimes$           & 64    & 35,840  &$\times$&\textbf{84.1} &\textbf{64.8}&       -  \\ 
$\phib_2\otimes$            & n/a  &  22,680  &        & 73.7         &  50.1       &       44.3\\  
$\phib_2\otimes$            & n/a   & 22,680  &$\times$& 79.9         &  60.5       &       51.9\\  
$\phib_3\otimes$            & n/a  & 265,680  &        & 72.5         &  53.5       &       -    \\ \hline  
\end{tabular}
\smallskip
\caption{Performance comparison with state of the art approaches. Results with the use
of full vector representation. \#C: size of codebook. \#dim: Number of components of each vector. Modulation
is performed with $N=3$ for all cases, except to $\phib_3$, where $N=1$. 
We do not use any re-ranking or spatial verification in any experiment. 
VLAD$\otimes$ achieves \textbf{87.2} on Holidays and 50.5 on Oxford5k with $\#C$=512 and oriented dense.
\label{tab:soa}}
\end{table}

\newcommand{\upd}[1]{\scriptsize (+#1)}

\begin{table}
\centering
\begin{tabular}{|@{\msp}l@{\msp}|@{\msp}r@{\msp}|@{\msp}c@{\msp\msp}l@{\msp\msp}l@{\msp}|} 
\hline
Method          & \#dim  & full dim  & dim$\rightarrow$1024 & dim$\rightarrow$128 \\ \hline \hline
VLAD              & 4,096   & 40.3  & 34.7          & 24.0 \\ 
VLAD$\otimes$   & 28,672  & 53.9  &\textbf{40.7}\,\upd{7.0}  & \textbf{27.5}\,\upd{3.5} \\  
\hline
Fisher            & 2,560   & 39.3  & 37.3          & 25.2 \\ 
Fisher$\otimes$ & 17,920  & 52.2  & 39.9 \upd{2.6}          & 26.5 \upd{1.3} \\   
\hline
$\phib_2$         & 3,240   & 35.8  & 31.1          & 20.4 \\   
$\phib_2\otimes$ & 22,680  & 51.9  & 37.7 \upd{6.6}           & 24.0 \upd{3.6} \\   \hline
\end{tabular}
\smallskip
\caption{Oxford105k: Performance comparison (mAP) after dimensionality reduction with PCA into 128 and 1024 components. The results with the full vector representation are with RN. Observe the consistent gain (in parentheses) brought by our approach for a \emph{fixed} output dimensionality of 1,024 or 128 components. 
\label{tab:pca}}
\vspace{-3ex}
\end{table}

We compare our approach, in particular, to CVLAD, as this work also intends to integrate the dominant orientation into a vector representation. We consistently apply 8 query rotations for both CVLAD and our method on Holidays dataset. Figure~\ref{fig:comparecvlad} shows the respective performance measured for
different codebooks. The proposed methods appear to consistently outperform CVLAD, both for the same codebook and for the same dimensionality. Noticeably, the modulated embedded monomial $\phib_2\otimes$  is on par with or better than CVLAD. 

We further conduct experiments using oriented dense~\cite{ZJG13} to compare VLAD$\otimes$ to CVLAD. 
They achieve 87.2 and 86.5 respectively, on Holidays with codebook of size 512. This score is significantly higher than the one reported in~\cite{ZJG13}. Corresponding scores on Oxford5k are 50.5 and 50.7, respectively. However, note that it is very costly to densely extract patches aligned with dominant orientation.

We also compare to other prior works and present results in Table~\ref{tab:soa} for Holidays, Oxford5k and Oxford105k. We outperform by a large margin the state of the art with full vector representations. Further, our approach is arguably compatible with these concurrent approaches, which may bring further improvement. Note that RN also boosts performance for VLAD and Fisher. In particular with a codebook of size 32, they achieve $50.0$ and $48.6$ respectively on Oxford5k.
Our scores on Holidays with Fisher$\otimes$ and RN are also competitive to those reported  by state-of-the-art methods based on large codebooks~\cite{TAJ13}. To our knowledge, this is the first time that a vector representation compatible with inner product attains such image search performance. 

On Oxford5k we do not evaluate multiple query rotations for our method. A simple way to enforce up-right objects for baseline methods is to use up-right features. Performance of VLAD with codebook of size 256 decreases from 51.3 to 49.4 by doing so, presumably because of small object rotations.

Finally, Table~\ref{tab:pca} reports the performance after dimensionality reduction to 128 or 1024 components.  The same set of local features and codebooks are used for all methods.  We observe a consistent improvement over the original encoding.

\vspace{-2ex}
\subsection{Timings}
\vspace{-1ex}
The image representation created by modulating the monomial embedding $\phib_2$ using $N=3$
takes on average 68\,ms for a typical image with 3,000 SIFT descriptors. The resulting aggregated vector representation has 22,680 components. 
The average query time using cosine similarity on Oxford5k is 44\,ms assuming no query rotation and
257\,ms with the use of 8 possible fixed rotations (with the naive strategy discussed in Section~\ref{sec:rotationinvariance}). The corresponding timings for Oxford105k and vectors reduced to 128 dimensions are 55\,ms and 134\,ms, respectively. Note, these timings are better than those achieved by a bag-of-words representation with a large vocabulary, for which the quantization typically takes above 1 second with an approximate nearest neighbor search algorithm like FLANN~\cite{ML09}. 
\vspace{-3ex}

\vspace{-1ex}
\section{Conclusion}
\vspace{-1ex}
Our modulation strategy integrates the dominant orientation directly in the coding stage. 
It is inspired by and builds upon recent works on explicit feature maps and kernel descriptors. Thanks to a generic formulation provided by match kernels, it is compatible with coding strategies such as Fisher vector or VLAD. Our experiments demonstrate that it gives a consistent gain compared to the original coding in all cases, even after dimensionality reduction. Interestingly, it is also very effective with a simple monomial kernel, offering competitive performance for image search with a coding stage not requiring any quantization. 

Whatever the coding stage that we use with our approach, the resulting representation is compared with inner product, which suggests that it is compliant with linear classifiers such as those considered in image classification.

\paragraph{Acknowledgments.}

This work was supported by ERC grant \textsc{Viamass} no. 336054 and ANR project Fire-ID. 

\bibliographystyle{ieeetr}
\bibliography{egbib}
\end{document}